\documentclass[journal]{IEEEtran}

\usepackage[T1]{fontenc}
\usepackage{amsmath,amssymb,amsthm}
\usepackage{booktabs}
\usepackage{graphicx}
\usepackage[colorlinks=true,linkcolor=blue,citecolor=blue,urlcolor=blue]{hyperref}
\usepackage[numbers,sort&compress]{natbib}
\usepackage{microtype}
\usepackage{array}
\usepackage{tabularx}
\usepackage{multirow}
\usepackage{xcolor}
\usepackage{enumitem}
\graphicspath{{figures/}}

\newcommand{\M}{\mathcal{M}}
\newcommand{\HO}{\textsc{Human\nobreakdash-Only}}
\newcommand{\CO}{\textsc{Copilot}}
\newcommand{\PE}{\textsc{Peer}}
\newcommand{\SU}{\textsc{Supervised}}
\newcommand{\AU}{\textsc{Autonomous}}
\newcommand{\HRS}{\mathrm{HRS}}
\newcommand{\tup}{\tau_{\uparrow}}
\newcommand{\tdown}{\tau_{\downarrow}}
\newcommand{\BASELINE}{\textsc{Baseline}}
\newcommand{\GOV}{\textsc{Governance}}
\newcommand{\READ}{\textsc{Readiness}}
\newcommand{\PLAT}{HAAS~Studio}

\title{The Handover Problem: Governing Autonomy Transitions in
       Human--AI Collaboration}

\author{Vicente Pelechano, Antoni Mestre, Manoli Albert, and Miriam Gil%
\thanks{V.~Pelechano, A.~Mestre, and M.~Albert are with VRAIN,
Universitat Polit\`{e}cnica de Val\`{e}ncia, Valencia, Spain
(e-mail: \{pele, amestre, malbert\}@vrain.upv.es).}%
\thanks{M.~Gil is with Universitat de Val\`{e}ncia, Valencia, Spain
(e-mail: miriam.gil@uv.es).}}

\begin{document}

\maketitle

% Disable em-dash substitution for repeated authors (IEEEtranN.bst)
\bstctlcite{IEEEexample:BSTcontrol}

% =============================================================
%  ABSTRACT
% =============================================================

\begin{abstract}
Human-machine systems rarely operate at a fixed level of AI autonomy.
As operators and AI systems collaborate over time, control must shift:
the AI can take on more responsibility when collaboration is stable,
maintain its current role when evidence is ambiguous, or return control
to the human when conditions deteriorate. Existing work on adaptive automation,
supervisory control, trust in automation, and deskilling explains parts
of this problem, but provides no auditable, multi-signal criterion for
governing when autonomy should change across multi-cycle workflows.

We formalise this challenge as the \emph{Handover Problem}: deciding, at
each operational cycle, whether to escalate, maintain, or revert AI
autonomy while keeping the process reversible, recoverable, and auditable.
We introduce the \emph{Handover Readiness Score} (HRS), a transparent
composite measure that integrates four signal dimensions --- operator
readiness, human--AI trust, learning stability, and operational
performance --- together with a hysteresis-based transition policy that
requires sustained positive evidence before increasing autonomy but
reverts promptly when conditions worsen.

Across software engineering and manufacturing domains, the HRS and hard
safety guards address complementary failure regimes: guards enforce
immediate corrective action when a single indicator breaches a critical
threshold; the HRS detects the slow, multi-signal erosion of operator
readiness that no individual guard can observe. The framework establishes
autonomy handover as a governance problem requiring explicit, composite,
and auditable criteria --- a conceptual and formal foundation that
adaptive automation research has not previously provided.
\end{abstract}


% =============================================================
%  1. INTRODUCTION
% =============================================================

\section{Introduction}
\label{sec:intro}

Autonomy is not a destination. In operational human--AI systems, the
central challenge is rarely whether to delegate tasks to an artificial
agent, but rather when to escalate delegation, how far, and --- equally
important --- when to take control back. A process-plant operator who has
supervised an increasingly autonomous control system for six months may
show stable metrics while quietly losing hands-on competence: no single
threshold was crossed, yet the operator's fallback capability has eroded.
A composite criterion sensitive to the \emph{joint} trajectory of
readiness, trust, learning stability, and operational performance would
have detected this drift and temporarily reverted to a more
operator-led mode, buying time to rebuild readiness before the skill gap
became consequential. Fixed-threshold guards alone cannot act here.

Several research traditions bear directly on this problem --- shared
autonomy, adaptive automation, trust in automation, and deskilling ---
yet none integrates these dynamics into a single temporal criterion for
governing escalation and reversion across multi-cycle workflows. This paper
proposes such a criterion and evaluates it via controlled simulation as a
governance mechanism for auditable, reversible, and recoverable autonomy
transitions.

We make four contributions:
\begin{enumerate}[leftmargin=*, label=\textbf{C\arabic*.}]
  \item \textbf{The Handover Problem}: a formal definition of governing
        transitions between collaboration modes, characterizing the tension
        between immediate performance and deferred deskilling costs, and
        specifying reversibility, recoverability, and auditability as
        requirements.
  \item \textbf{The Handover Readiness Score (HRS)}: a composite index
        aggregating human readiness, relational trust, learning stability,
        and operational performance into a single interpretable scalar.
        We additionally analyse a calibrated trust extension that penalises
        trust--performance mismatch, addressing overtrust-driven false
        escalations in controlled experiments.
  \item \textbf{A reversible hysteresis policy} that encodes asymmetric
        caution: moving to higher autonomy requires sustained evidence of
        readiness across consecutive cycles, while reverting to a safer mode
        is never delayed. The design reflects the principle that granting
        autonomy too early is a harder mistake to recover from than
        temporarily withholding it.
  \item \textbf{Complementarity of guards and HRS}: guards and the HRS
        address structurally distinct failure regimes --- guards respond to
        abrupt threshold crossings; the HRS detects slow, multi-signal erosion
        that no individual guard observes. This separation is confirmed across
        software engineering and manufacturing domains.
\end{enumerate}

The primary contribution is a formally specified governance criterion:
each component is individually motivated, and the framework is shown
to produce effects distinguishable from simpler guard-based alternatives.
The framework establishes autonomy handover as a governance problem
requiring explicit, composite, and auditable criteria --- a conceptual
and formal foundation that adaptive automation research has not
previously provided.

% =============================================================
%  2. RELATED WORK
% =============================================================

\section{Related Work}
\label{sec:related}

\textbf{Shared autonomy and adjustable automation.}
\citet{sheridan1978human} proposed a 10-level taxonomy from fully manual
to fully automated operation; \citet{parasuraman2000model} extended this
into a two-dimensional model distinguishing automation types from levels
of human involvement. Sarter et al.~\cite{sarter1997automation} documented automation
surprises --- transitions without adequate human awareness --- as a
primary source of operational risk. Inagaki~\cite{inagaki2003adaptive}
distinguished sharing (concurrent execution) from trading (sequential
handoff), a contrast central to the governance problem studied here;
Flemisch et al.~\cite{flemisch2003h} placed sharing and trading on a
common cooperative-control spectrum. Automation level and adaptive
automation further shape performance, situation awareness, and workload
\citep{kaber2004effects}.
The HRS most closely resembles a \emph{hybrid trigger} in
\citet{parasuraman1997humans}'s taxonomy --- one combining multiple
signal types --- but extends it to govern both escalation and reversion
with explicit hysteresis and component-level auditability.

\textbf{Trust in automation and human oversight.}
\citet{lee2004trust} distinguish appropriate reliance from overtrust
(automation bias) and undertrust (disuse); trust evolves dynamically through experience
\citep{muir1994trust,hoff2015trust} and is more brittle than interpersonal
trust after failure \citep{hancock2011meta}.
Miscalibrated trust has documented consequences: humans preferentially
accept automated recommendations even when contradicted by evidence
\citep{skitka1999does,goddard2012automation}.
\citet{shneiderman2020human} and \citet{cummings2014man} argue that
accountable oversight mechanisms --- not merely capable AI --- are the
operative model for consequential deployments; \citet{christoffersen2002make}
show that making joint cognitive work visible to both operators and designers
is a prerequisite for effective oversight. These findings motivate
treating reliance as a state to be actively governed.

\textbf{Human capital erosion and deskilling.}
Bainbridge's~\cite{bainbridge1983ironies} foundational analysis
identified the central paradox: reliable automation removes practice
opportunities for the skills most needed when automation fails.
\citet{endsley1995outofloop} documented empirically that reduced control
involvement leads to lower situation awareness and slower reversion
responses; \citet{endsley2017from} concluded that full autonomy
systematically undermines skill retention and meaningful oversight.
Aviation offers the most extensively documented domain
\citep{sarter1994pilot,parasuraman1993performance},
and the concern has migrated to knowledge work with generative AI
assistants \citep{barke2023grounded,heudel2026deskilling}.

\textbf{Autonomy governance in agentic AI systems.}
The rise of LLM-based agents reframes handover as an active engineering
concern: AI systems now pursue multi-step goals with partial autonomy,
requiring humans to intervene selectively
\citep{shneiderman2020human,cummings2014man}. Recent work on
human-in-the-loop oversight \citep{amershi2019guidelines} and on the
cognitive costs of interruption-based control
\citep{parasuraman2000model} shows that neither ``always ask'' nor
``always act'' is satisfactory. The EU AI Act increases the need for
risk management, human oversight, and traceability in high-risk AI
settings, creating a regulatory demand for auditable governance criteria
\citep{euaiact2024}.
The Handover Problem addresses this demand by formalising escalation and
reversion as first-class governance actions with explicit, component-traceable
criteria rather than ad hoc human overrides.

\textbf{The intersection gap.}
Shared autonomy provides the mode spectrum, trust a key dynamic signal,
and deskilling the forward-looking risk that should govern reversion.
What is absent is a framework using all three together as a temporal
governance criterion across multi-cycle workflows.
\citet{wickens2002multiple}'s multiple-resources framework grounds this:
readiness can erode along independent dimensions --- high skill does not
compensate deteriorating trust, nor does a stable operational period
offset deskilling accumulation. The Handover Problem formalises this gap.

% =============================================================
%  3. THE HANDOVER READINESS MODEL
%     (merges: The Handover Problem + Platform + Transition Model)
% =============================================================

\section{The Handover Readiness Model}
\label{sec:model}

\subsection{Mode Spectrum and Handover Problem}
\label{sec:model:problem}

An \textit{operational cycle} $t \in \{1,\ldots,T\}$ is a bounded
collaboration period. The collaboration mode spectrum is:
\begin{equation}
\begin{gathered}
\M = \{\,\HO \prec \CO \prec \PE\\
      \prec \SU \prec \AU\,\}.
\end{gathered}
\end{equation}
where $\prec$ denotes increasing AI autonomy. Table~\ref{tab:modes}
summarises the five modes. In \CO{} the human leads while the AI
assists; in \PE{} both agents execute independently; in \SU{} the AI
leads while the human validates.

\begin{table}[ht]
\centering
\caption{The five collaboration modes ordered by increasing AI autonomy.
Intermediate modes (\CO{}, \PE{}, \SU{}) are \textit{shared}: both
agents execute portions of each task.}
\label{tab:modes}
\small
\setlength{\tabcolsep}{3pt}
\renewcommand{\arraystretch}{1.15}
\begin{tabular}{@{}
  l
  >{\raggedright\arraybackslash}p{0.37\linewidth}
  >{\raggedright\arraybackslash}p{0.32\linewidth}
  @{}}
\toprule
\textbf{Mode} & \textbf{Human role} & \textbf{AI role} \\
\midrule
\HO{}        & Full execution      & None \\
\CO{}        & Leads, decides      & Assists \\
\PE{}        & Executes autonomously & Executes autonomously \\
\SU{}        & Validates outputs   & Leads, executes \\
\AU{}        & None                & Full execution \\
\bottomrule
\end{tabular}
\end{table}

\paragraph*{The Two-Horizon Tension}
Escalating AI autonomy confers immediate benefits: faster execution, lower
cost, consistent output quality. The deferred cost is more complex. Operating
at high autonomy for many consecutive cycles accumulates a deskilling
liability --- the human's fallback capability erodes through disuse
\citep{bainbridge1983ironies,endsley1995outofloop}. Trust may also
miscalibrate: extended AI success produces overtrust that impairs detection
of AI failures \citep{goddard2012automation}. A policy that maximises
immediate performance escalates aggressively, accumulating fragility; a
policy that minimises deskilling risk under-utilises AI capability.
The Handover Problem arises because these forces pull in opposite
directions, and resolving the tension requires a principled composite
criterion --- not a single threshold.

The \textbf{Handover Problem} is: design a policy $\pi: S \to
\{\textsc{escalate},\allowbreak\textsc{maintain},\allowbreak\textsc{revert}\}$
satisfying: (i)~\textbf{completeness} (a decision at every cycle);
(ii)~\textbf{reversibility} (reversion is a first-class action, not
reserved for failure); (iii)~\textbf{auditability} (every decision
traces to observable signals with explicit weights);
(iv)~\textbf{balance} (neither maximising performance at deskilling
cost nor sacrificing AI capability entirely); and
(v)~\textbf{recoverability} (the policy preserves the operator's
capacity to absorb reverted control --- reversibility guarantees control
is returnable; recoverability ensures returning it remains useful).
These requirements distinguish the framework from reactive threshold
rules, which satisfy~(i) but fail (ii)--(v).

\subsection{The Handover Readiness Score}
\label{sec:model:hrs}

The Handover Readiness Score at cycle $t$ is:
\begin{equation}
\HRS_t = w_H \cdot H_t + w_R \cdot R_t + w_L \cdot L_t + w_S \cdot S_t,
\label{eq:hrs}
\end{equation}
where $w_H + w_R + w_L + w_S = 1$ and $\HRS_t \in [0,1]$. The four
components correspond to four distinct reasons the answer to ``is the
system ready to escalate?'' may be ``not yet'': the human may be unable
to function as an effective fallback ($H_t$), the relational basis for
delegation may be weak ($R_t$), the learning trajectory may be unstable
($L_t$), or the current mode may lack demonstrated equilibrium ($S_t$).

\paragraph*{Human readiness $H_t$}
\begin{equation}
H_t = (1 - f_t) \cdot \sigma_t \cdot (1 - d_t),
\label{eq:H}
\end{equation}
combining fatigue ($f_t$), skill level ($\sigma_t$), and deskilling
exposure ($d_t$). The multiplicative form is deliberate: any severely
degraded factor collapses the product regardless of the others ---
readiness is the coexistence of prerequisites, not an average.
$H_t$ is the formal carrier of recoverability: it is the quantity
through which the policy observes whether returned control would still be
usable. Weight $w_H = 0.35$ is highest.

\paragraph*{Relational trust $R_t$}
$R_t = \tau_t$, the normalised learned trust signal. Trust builds
gradually through consistent AI performance and degrades abruptly after
failures \citep{muir1994trust}. A hard guard at $\tau_t < 0.25$ provides
a safety floor. Note: setting $R_t = \tau_t$ means higher trust always
raises the composite, which can reward escalation under overtrust ---
a substantive limitation discussed in Section~\ref{sec:discussion}.

\paragraph*{Learning stability $L_t$}
\begin{equation}
L_t = \max\!\left(0,\; 1 - \frac{\mathrm{Var}(\sigma_{t-k:t})}{\sigma^2_{\max}}\right),
\end{equation}
asking whether the skill trajectory has settled, not just whether the
current level is adequate. Window $k=5$ cycles; $\sigma^2_{\max}=0.04$
(i.e.\ $\sigma_{\max}=0.2$, held constant per domain to prevent baseline
drift). Weight $w_L = 0.20$.

\paragraph*{Operational stability $S_t$}
\begin{equation}
S_t = \mathrm{clip}\!\left(\frac{\kappa_t}{\kappa_0},0,1\right)
      \cdot
      \frac{\ln(1+\min(c_t,k_{\max}))}{\ln(1+k_{\max})},
\end{equation}
combining current KPI relative to mode baseline $\kappa_0=0.015$ with
logarithmic tenure $c_t$ in the current mode (cap $k_{\max}=10$). Weight
$w_S = 0.20$. Note: immediately after any mode transition $c_t=0$, so
$\ln(1+0)=0$ collapses $S_t$ to zero regardless of KPI. This is
deliberate --- the new mode has no demonstrated tenure and should not
contribute stability evidence --- but the resulting transient $\HRS_t$
dip is bounded: it cannot drive reversion unless $\HRS_t$ without the
$S_t$ contribution already sits near $\tdown$, i.e.\ unless the other
three components are also weak. Guard conditions are unaffected by $S_t$.

\paragraph*{Default weights: $w_H=0.35$, $w_R=0.25$, $w_L=0.20$,
$w_S=0.20$}
These are design parameters encoding organisational priorities, not
fitted coefficients, making each transition decision auditable by
component. Table~\ref{tab:signals} lists the observable proxies used
to instantiate each component in the simulation and in candidate
deployments.

\begin{table}[ht]
\centering
\caption{Observable proxies for each HRS component and acquisition
method for real deployments. All signals are derivable from telemetry
already present in most operational human--AI systems.}
\label{tab:signals}
\small
\renewcommand{\arraystretch}{1.25}
\begin{tabular}{@{} l p{2.0cm} p{2.2cm} @{}}
\toprule
\textbf{Component} & \textbf{Field proxy} & \textbf{Acquisition source} \\
\midrule
$H_t$ — fatigue $f_t$
  & Session length, input latency, error clustering
  & System telemetry; interaction logs \\
$H_t$ — skill $\sigma_t$
  & Task-type success rate, competency assessments
  & QA/audit logs; HR system \\
$H_t$ — deskilling $d_t$
  & Time since last unaided execution of task type
  & Mode-history ledger \\
$R_t$ — trust $\tau_t$
  & Override frequency, confirmation delay, AI-accept rate
  & Interaction audit trail \\
$L_t$ — learning stab.
  & Rolling variance of task-quality over $k$-cycle window
  & QA logs (computed) \\
$S_t$ — op. stability
  & KPI vs.\ mode baseline; cycles in current mode
  & Operational dashboard; scheduler \\
\midrule
Guards ($f_t$, $d_t$, $\tau_t$)
  & Same proxies as above; hard-threshold check
  & Real-time policy engine \\
\bottomrule
\end{tabular}
\end{table}

\subsection{Hysteresis Policy and Hard Guards}
\label{sec:model:policy}

The policy translates the HRS scalar into a transition decision ---
escalate, maintain, or revert --- at each cycle. A single threshold
would cause oscillation near the boundary, as small HRS fluctuations
trigger repeated mode changes. Instead, two thresholds $\tup > \tdown$
define three zones: escalation ($\HRS_t \geq \tup$), maintenance band
($\tdown < \HRS_t < \tup$), and reversion ($\HRS_t \leq \tdown$). Let $n^+_t$ count consecutive cycles
with $\HRS_s \geq \tup$, resetting on any dip. The policy is:
\begin{equation}
D_t = \begin{cases}
\textsc{revert}   & \text{if } \mathrm{guard}(s_t) = \top \\[3pt]
\textsc{escalate} & \text{else if } n^+_t \geq k_{\uparrow} \\[3pt]
\textsc{revert}   & \text{else if } \HRS_t \leq \tdown \\[3pt]
\textsc{maintain} & \text{otherwise.}
\end{cases}
\end{equation}
Default parameters: $\tup=0.72$, $\tdown=0.40$, $k_{\uparrow}=3$. These
defaults were derived from empirical calibration on the evaluation
platform and validated across experimental conditions in
Section~\ref{sec:results}; deployment requires domain-specific
recalibration.
All transitions are single-step. The asymmetry is deliberate: escalation
is a commitment that exposes the system to deskilling risk; reversion is
a correction. Sustained evidence is required to reduce human involvement;
reversion is never delayed.

Hard guards override the HRS and force immediate reversion:
\begin{equation}
\begin{aligned}
\mathrm{guard}(s_t)=\top \iff{}&
  (f_t>0.85)\lor(\tau_t<0.25)\\
&{}\lor(d_t>0.60)\lor(\mathrm{fail\_streak}_t\geq\theta_f).
\end{aligned}
\label{eq:guard}
\end{equation}
The default $\theta_f=3$ (three consecutive task failures) is domain-configurable.
Each condition marks a qualitatively distinct failure mode that the
composite must not obscure. \emph{Fatigue} ($f_t>0.85$) signals acute
incapacity: at this level the operator cannot reliably execute fallback
tasks regardless of skill level or trust \citep{wickens2002multiple}.
\emph{Trust} ($\tau_t<0.25$) signals relational breakdown: below this
floor the operator has effectively disengaged from AI outputs, making
delegation unsafe irrespective of AI capability \citep{lee2004trust}.
\emph{Deskilling} ($d_t>0.60$) signals a point of uncertain
recoverability: sustained exposure beyond this threshold approaches the
irreversible skill-gap regime documented in automation-heavy operations
\citep{endsley1995outofloop,bainbridge1983ironies}.
\emph{Fail streak} ($\mathrm{fail\_streak}_t \geq \theta_f$) signals
sustained output failure: three consecutive task failures indicate a
systemic collaboration breakdown, not random noise \citep{parasuraman1997humans}.
Guards mark states the governance policy refuses to ``average away'':
these are stop conditions, not trade-offs. Two guard conditions
($f_t$, $d_t$) act directly on erosion factors of $H_t$, functioning as
backstops for recoverability before fallback capability degrades
irrecoverably.

\paragraph*{Illustrative mechanism}
Suppose, with $\tdown=0.55$ (as calibrated for the H3 scenario in
Section~\ref{sec:results:shocks}), skill erodes from $0.74$ to $0.57$
and deskilling exposure grows from $0.26$ to $0.43$ under sustained
high-autonomy operation. Each shift is modest in isolation, yet their
product inside $H_t$ (with $f=0.01$) collapses human readiness to
$H=0.322$, giving $\HRS = 0.35(0.322)+0.25(0.93)+0.20(0.99)+0.20(0.00)
= 0.544 \leq \tdown$. The composite thus crosses the reversion threshold
that no individual guard can observe.

\begin{figure}[t]
\centering
\includegraphics[width=0.85\linewidth]{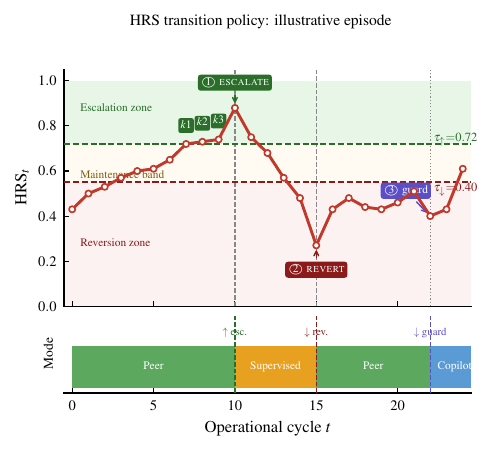}
\caption{HRS-governed transition policy on a hypothetical 25-cycle
episode.
\textit{Top}: $\HRS_t$ trajectory with zone bands (escalation, maintenance,
reversion) and escalation counter ($k{=}1,2,3$ at cycles 7--9).
\textit{Bottom}: resulting collaboration mode strip.
Events: \textcircled{1}~escalation after $k_{\uparrow}{=}3$ consecutive
cycles above $\tup$ (cycle~10); \textcircled{2}~HRS-triggered reversion
when $\HRS_t$ drops below $\tdown$ on a trust shock (cycle~15);
\textcircled{3}~hard-guard reversion on a fatigue spike (cycle~22),
overriding the composite score.}
\label{fig:transition_model}
\end{figure}

Figure~\ref{fig:transition_model} traces a hypothetical 25-cycle episode
illustrating the three key policy behaviours: \textcircled{1}~escalation
fires after $k_{\uparrow}{=}3$ consecutive cycles above $\tup$ (cycle~10);
\textcircled{2}~an AI error collapses $\HRS_t$ into the reversion zone and
the policy acts immediately (cycle~15); \textcircled{3}~a fatigue spike
triggers the hard guard, overriding $\HRS_t$ (cycle~22). Cycles~1--6 and
16--21 illustrate the maintenance band: the system holds the current mode
while $\HRS_t$ is neither low enough to revert nor high enough (for long
enough) to escalate.

Empirical validation of the full model --- ablations, severity sweeps,
sensitivity maps, and healthcare boundary results --- appears in the
companion technical report~\cite{pelechano2026thms_supp}, Sections~S1--S5.

% =============================================================
%  4. METHODOLOGY
% =============================================================

\section{Methodology}
\label{sec:method}

\textbf{Experimental platform.}
\PLAT{} is a computational simulation platform for human--AI
collaboration research, built on the HAAS policy-aware task-allocation
framework \citep{pelechano2026haasframework,pelechano2026haasstudiotoolsimulating}.
Each model component is a formal instantiation of an established construct:
fatigue follows workload-capacity dynamics \citep{wickens2002multiple};
trust evolves through gradual reinforcement and abrupt failure-degradation
\citep{muir1994trust,lee2004trust}; deskilling accrues proportionally to
AI-exposure share \citep{bainbridge1983ironies,endsley1995outofloop}.
Parameters ($\kappa_0=0.015$, $\sigma^2_{\max}=0.04$, $\theta_f=3$) are
calibrated to qualitative benchmarks in the literature, not fitted to the
outcomes reported here. The evaluation therefore tests whether the HRS
behaves coherently \emph{given that the underlying dynamics follow the
theoretically assumed patterns} --- the necessary computational precondition
before field deployment.

Each episode of $T{=}100$ cycles generates domain-specific subtasks; a
\textsc{LinUCB} contextual bandit \citep{li2010contextual} proposes a
human--AI collaboration mode, which the policy engine may override via
caps, guards, or the HRS. The human-agent model maintains state variables for fatigue,
skill level, deskilling exposure, and trust; the AI-agent model updates
its capability based on the learning trajectory. Outcomes are recorded
and fed back as input for the next cycle. \textsc{LinUCB} is chosen
because it conditions on the same state signals the HRS observes,
creating a demanding baseline: any risk the HRS corrects cannot be
attributed to allocator ignorance. Three domains are modelled: software
engineering (moderate deskilling decay, no regulatory ceiling),
manufacturing (faster physical-skill decay), and healthcare (fastest
judgment-skill decay, hard regulatory ceiling at \SU{}).

Three experimental conditions isolate the contribution of each
governance layer (Table~\ref{tab:conditions}): \BASELINE{} (LinUCB
allocator, no governance), \GOV{} (cap/guard architecture, no HRS), and
\READ{} (cap/guard architecture plus the HRS policy). \GOV{} is
reactive: it acts only when a single-signal threshold is violated.
\READ{} adds a proactive, four-component readiness trajectory and
governs escalation, maintenance, and reversion through the same
auditable criterion. The key contrast is \READ{} vs.\ \GOV{}: both
share the same cap/guard architecture, so any difference isolates the
HRS's specific contribution.

\begin{table}[ht]
\centering
\caption{Architectural differences between experimental conditions.}
\label{tab:conditions}
\scriptsize
\setlength{\tabcolsep}{2pt}
\renewcommand{\arraystretch}{1.15}
\begin{tabular}{@{}
  >{\raggedright\arraybackslash}p{0.22\linewidth}
  >{\raggedright\arraybackslash}p{0.17\linewidth}
  >{\raggedright\arraybackslash}p{0.24\linewidth}
  >{\raggedright\arraybackslash}p{0.28\linewidth}
  @{}}
\toprule
\textbf{Dimension} & \BASELINE{} & \GOV{} & \READ{} \\
\midrule
Governance & None & Caps + guards & Caps + guards + HRS \\
Direction  & --- & Reactive & Proactive \\
Signals    & --- & Single-signal & Composite (4 dim.) \\
Reversibility & --- & Guard-fire only & Escalate/maint./revert \\
Hysteresis & --- & None & $k_\uparrow$ cycles; immediate revert \\
\bottomrule
\end{tabular}
\end{table}

The primary domain is software engineering ($T=100$ cycles, 50 random
seeds). Manufacturing serves as a replication domain. Healthcare
(radiology screening, regulatory ceiling at \SU{}) maps the governance
boundary.

\textbf{Hypotheses.}
The four hypotheses follow a progressive logic.
H1 and H2 characterise the governance layer in steady-state operation,
where the cap/guard architecture alone is expected to be sufficient.
H3 identifies the HRS's distinctive contribution: detecting slow,
multi-signal deterioration that no single guard can observe.
H4 confirms that the governance system is robust across a meaningful
region of the parameter space.

\begin{description}[leftmargin=2.5em, labelwidth=2em, labelsep=0.5em,
                    itemsep=2pt, parsep=0pt]
\item[\textbf{H1}] \textit{(Mode redistribution)}: Governed conditions
  reduce high-autonomy concentration relative to \BASELINE{}; the HRS
  adds no separate mode signature in steady state, since both governed
  conditions share the same cap/guard layer.

\item[\textbf{H2}] \textit{(Sustainability)}: The cap/guard layer
  reduces deskilling risk relative to \BASELINE{}; the HRS adds no
  further steady-state gain, as its contribution operates through
  deterioration detection rather than mode redistribution.

\item[\textbf{H3}] \textit{(Complementarity)}: \READ{} achieves higher
  reversion coverage under gradual sub-threshold deterioration than
  \GOV{}/\BASELINE{}, which structurally cannot act in the guard-silent
  regime.

\item[\textbf{H4}] \textit{(Stability)}: A broad stable region $\Omega$
  (handover accuracy $\geq 0.75$) spans a wide range of threshold
  settings, with the default configuration interior to it.
\end{description}

\textbf{Deterioration protocol.}
Protocol parameters are calibrated relative to the simulation's
domain-specific dynamics. Under normal operation, software-domain skill
decays at 0.002--0.004 per cycle per dimension, representing gradual
attrition under routine AI exposure
\citep{bainbridge1983ironies,endsley1995outofloop}. Starting at cycle
$t_0=15$, the H3 protocol sets $\Delta\sigma=0.010$/cycle (floor 0.42)
--- roughly 2.5--5$\times$ baseline, a period of heightened degradation
consistent with more rapid skill erosion documented in aviation and
knowledge work \citep{casner2013effectiveness,endsley2017from}. Fatigue
accrues at $\Delta f=0.012$/cycle (cap 0.80)
\citep{wickens2002multiple}. The floor and cap keep the scenario in a
recoverable, non-emergency range, deliberately below guard thresholds
($f\!>\!0.85$, $d\!>\!0.60$), so that $\HRS_t$ is the only mechanism
capable of detecting the drift over 40 cycles.

The reversion threshold is recalibrated to $\tdown=0.55$, matching the
lower boundary of the observed HRS operating range in the software
domain. This value exceeds the default $\tdown=0.40$, making reversion
more conservative: the HRS triggers a return to human-led modes at a
higher readiness floor. It remains within the stable region $\Omega$
(Section~\ref{sec:results:sensitivity}); it is a deliberate stress-test
choice, not an operational default.
The primary estimand is \emph{reversion coverage}: the fraction of seeds
achieving majority human-led allocation (\CO{} or \HO{}) during the
erosion window.

\textbf{Primary metrics.}
Three metric classes operationalise the four hypotheses:
\begin{description}[leftmargin=2.5em, labelwidth=2em, labelsep=0.5em,
                    itemsep=2pt, parsep=0pt]
\item[\textit{Sustainability} (H1,\,H2)] \textit{deskilling\_risk}
  $= (\sigma_0-\sigma_T)/\sigma_0$ quantifies relative skill loss from
  episode start ($\sigma_0$) to end ($\sigma_T$).
  \textit{responsible\_score} (RS, Eq.~\ref{eq:responsible}) combines
  performance balance ($PB$), shared-mode fraction ($SM$), human
  retention ($HR$), and autonomy restraint ($AR=1-\text{AU\%}$) ---
  design weights, not fitted coefficients:
\begin{equation}
RS = 0.40\,PB + 0.20\,SM + 0.25\,HR + 0.15\,AR.
\label{eq:responsible}
\end{equation}
\item[\textit{Deterioration} (H3)] Reversion coverage (fraction of 50
  seeds reverting) and median reversal latency among reverting seeds.
\item[\textit{Sensitivity} (H4)] Handover accuracy: the fraction of
  governance decisions confirmed by subsequent quality and skill
  trajectories.
\end{description}
Extended severity sweeps appear in~\cite{pelechano2026thms_supp}, Section~S2.

% =============================================================
%  5. RESULTS
% =============================================================

\section{Results}
\label{sec:results}

All results are from computational simulation on \PLAT{}; no human
participants were involved. Values are means across 50 seeds unless
stated. Reversion rates include Wilson 95\% CIs.

The results fall into two parts. \emph{First}, in normal operation,
the cap/guard layer protects human sustainability at no cost to
performance --- \READ{} performs comparably to \GOV{} in steady state,
consistent with H1 and H2. \emph{Second}, the HRS
makes its distinctive contribution under gradual deterioration: \READ{}
covers the guard-silent regime where \GOV{} and \BASELINE{} structurally
cannot act. Table~\ref{tab:hyp_outcomes} summarises outcomes for all four hypotheses.

\begin{table}[ht]
\centering
\caption{Hypothesis outcomes summary. All evidence from computational
simulation. ``Directional'' denotes small absolute differences from which
no statistical claim is drawn.}
\label{tab:hyp_outcomes}
\small
\renewcommand{\arraystretch}{1.2}
\begin{tabular}{@{} l p{3.7cm} l @{}}
\toprule
 & \textbf{Prediction} & \textbf{Outcome} \\
\midrule
\textbf{H1} & Governed conditions cut \AU{} share; \GOV{}/\READ{} coincide
  in steady state.
  & Supported \\
\textbf{H2} & Cap/guard layer reduces deskilling; HRS adds no
  further steady-state gain.
  & Consistent (directional) \\
\textbf{H3} & \READ{} reverts under gradual deterioration where
  \GOV{}/\BASELINE{} cannot.
  & Supported (50/50 vs.\ 0/50) \\
\textbf{H4} & A broad stable region $\Omega$ exists; defaults sit
  inside it.
  & Supported (91\% of grid) \\
\bottomrule
\end{tabular}
\end{table}

\subsection{Steady-State Performance and Sustainability (H1, H2)}
\label{sec:results:tradeoff}

\begin{figure}[t]
\centering
\includegraphics[width=\linewidth]{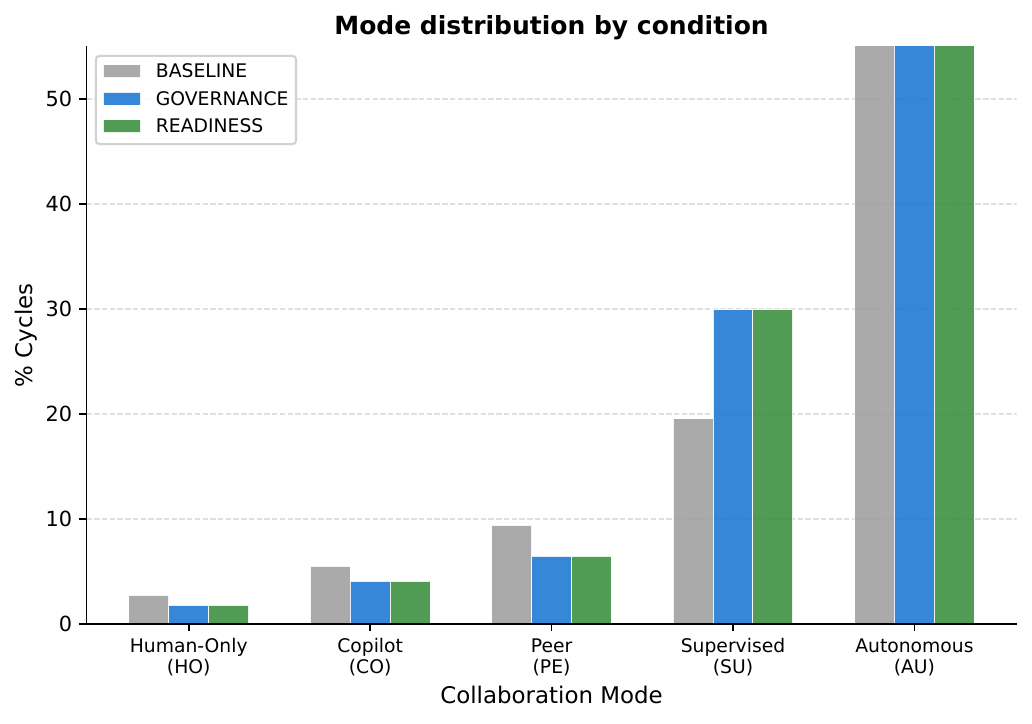}
\caption{Mode distribution across conditions (software domain, 50 seeds;
\textit{computational simulation}). Both governed conditions shift
allocation from \AU{} into \SU{} relative to \BASELINE{}. \GOV{} and
\READ{} are nearly indistinguishable in steady state (H1 confirmed):
the cap/guard layer, not the HRS, drives the redistribution.}
\label{fig:mode_dist}
\end{figure}

Table~\ref{tab:performance} reports steady-state metrics. Both governed
conditions reduce \AU{} share from 63\% (\BASELINE{}) to 42\%,
redirecting freed cycles into \SU{} (19\%~$\to$~29\%), where the AI
leads but the human validates. \GOV{} and \READ{} produce identical mode
distributions, exactly as H1 predicts (Fig.~\ref{fig:mode_dist}): they
share the same guard architecture, so the HRS adds no separate
steady-state footprint.

Average task quality is identical across all conditions (0.533--0.534),
confirming governance does not impair throughput. Deskilling risk
becomes measurable at $T=100$: \BASELINE{} accumulates 6.0\% skill
degradation; \GOV{} reduces this to 5.8\%, consistent with H2 (the
guard layer reduces deskilling; the HRS adds no further steady-state
gain). The mean \AU{} run under \READ{} (7.90 cycles) is shorter than
\GOV{} (8.07) and \BASELINE{} (8.09), reflecting the HRS reversion gate
cutting runs short when $\HRS_t$ falls below $\tdown$ mid-run --- but
the effect is too small to change $T=100$ deskilling accumulation
meaningfully. Manufacturing replicates the pattern with larger absolute
deskilling improvements (Table~\ref{tab:performance}, lower block),
reflecting that domain's faster physical-skill decay.

\begin{table}[ht]
\centering
\caption{Steady-state metrics, $T=100$ cycles, means over 50 seeds.
Deskilling risk $= (\sigma_0-\sigma_T)/\sigma_0$.}
\label{tab:performance}
\small

\renewcommand{\arraystretch}{1.2}
\begin{tabular}{@{} l c c c @{}}
\toprule
\textbf{Metric} & \textbf{Base} & \textbf{Gov} & \textbf{Read} \\
\midrule
\multicolumn{4}{l}{\textit{Software engineering (primary)}} \\
\quad \AU{} share (\%)      & $63$    & $\mathbf{42}$    & $\mathbf{42}$ \\
\quad \SU{} share (\%)      & $19$    & $\mathbf{29}$    & $\mathbf{29}$ \\
\quad avg\_quality          & $0.533$ & $0.533$          & $0.534$ \\
\quad deskilling\_risk      & $0.060$ & $\mathbf{0.058}$ & $0.061$ \\
\quad responsible\_score    & $0.315$ & $\mathbf{0.326}$ & $0.324$ \\
\quad mean \AU{} run (cyc.) & $8.09$  & $8.07$           & $\mathbf{7.90}$ \\
\midrule
\multicolumn{4}{l}{\textit{Manufacturing (replication)}} \\
\quad deskilling\_risk      & $0.065$ & $\mathbf{0.051}$ & $0.052$ \\
\quad responsible\_score    & $0.330$ & $\mathbf{0.372}$ & $\mathbf{0.372}$ \\
\bottomrule
\end{tabular}
\end{table}

\subsection{Gradual Deterioration Detection (H3)}
\label{sec:results:shocks}

\begin{figure}[t]
\centering
\includegraphics[width=\linewidth]{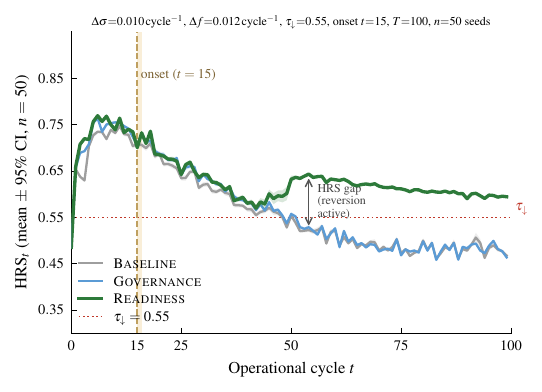}
\caption{HRS trajectory under gradual sub-threshold deterioration
(mean $\pm$ 95\% CI, $n=50$ seeds, software domain). All three
conditions decline together until the first composite crossing: \READ{}
crosses $\tau_{\downarrow}=0.55$ at cycle $\sim\!33$ ($\approx18.5$
cycles after onset at $t_0=15$), no individual guard fires because no
single threshold is crossed, and the policy activates reversion.
The HRS recovers to ${\approx}\,0.63$; the separation gap between
\READ{} and the other conditions is visible from cycle $\sim\!33$
onward. \BASELINE{} and \GOV{} continue falling below
$\tau_{\downarrow}$ and remain there: neither has a mechanism to
respond to the composite crossing. This explains the $50/50$
vs.\ $0/50$ reversion contrast (\textit{computational simulation}).}
\label{fig:reversal}
\end{figure}

This is where the HRS makes its distinctive contribution.
Table~\ref{tab:shocks} reports deterioration-response metrics.
In the software domain, \BASELINE{} and \GOV{} revert in 0/50 seeds:
with no threshold crossed, neither has a signal to act on. \READ{}
reverts in all 50/50 seeds (median latency 18.5 cycles; Wilson CI
[0.93,1.00] vs.\ [0.00,0.07] for the others). The contrast is
structural: the HRS fills the coverage gap that guards leave,
demonstrating \emph{complementarity}. The mechanism is direct: as skill
erodes, $H_t$ declines ($0.56\to0.22$), pulling $\HRS_t$ below
$\tdown=0.55$ around cycle $\sim\!33$ ($\approx18.5$ cycles after onset;
Fig.~\ref{fig:reversal}), applying a binding \CO{} cap without any single
guard firing. The result replicates in manufacturing (50/50, median
latency 22.0 cycles). These figures hold for the gradual sub-threshold
protocol specifically; secondary stress tests in~\cite{pelechano2026thms_supp}, Sections~S1--S3 extend the analysis to direct signal-injection
shocks.

\begin{table}[ht]
\centering
\caption{Deterioration-response metrics, gradual sub-threshold erosion
over cycles [15,55], $T=100$, 50 seeds. $^\ddagger$Median latency
(cycles from erosion onset) among reverting seeds; ``---'' when none
reverted. Wilson 95\% CI in brackets.}
\label{tab:shocks}
\scriptsize
\setlength{\tabcolsep}{2pt}
\renewcommand{\arraystretch}{1.18}
\begin{tabular}{@{}
  >{\raggedright\arraybackslash}p{0.36\linewidth}
  >{\centering\arraybackslash}p{0.18\linewidth}
  >{\centering\arraybackslash}p{0.18\linewidth}
  >{\centering\arraybackslash}p{0.20\linewidth}
  @{}}
\toprule
\textbf{Metric} & \textbf{Base} & \textbf{Gov} & \textbf{Read} \\
\midrule
\multicolumn{4}{l}{\textit{Software engineering (primary)}} \\
\quad reversion rate
  & $0/50$\newline{\scriptsize[0.00,0.07]}
  & $0/50$\newline{\scriptsize[0.00,0.07]}
  & $\mathbf{50/50}$\newline{\scriptsize[0.93,1.00]} \\
\quad latency$^\ddagger$ (cyc.) & --- & --- & $\mathbf{18.5}$ \\
\midrule
\multicolumn{4}{l}{\textit{Manufacturing (replication)}} \\
\quad reversion rate
  & $0/50$\newline{\scriptsize[0.00,0.07]}
  & $0/50$\newline{\scriptsize[0.00,0.07]}
  & $\mathbf{50/50}$\newline{\scriptsize[0.93,1.00]} \\
\quad latency$^\ddagger$ (cyc.) & --- & --- & $\mathbf{22.0}$ \\
\bottomrule
\end{tabular}
\end{table}

\textbf{Non-trivial comparators.}
A natural objection is that the $0/50$ vs.\ $50/50$ contrast is
tautological: \READ{} has a composite-linked reverting cap; the other
conditions do not. Table~\ref{tab:h3comparators} isolates which element
drives the advantage with two cap-equipped comparators:
(i)~$H_t$\textbf{-only}, a cap driven by the human-condition component
alone, and (ii)~\textbf{no-hysteresis}, the full composite with a single
threshold ($\tup=\tdown$, $k_{\uparrow}=1$). Each variant runs under
both deterioration and a no-deterioration control; reversion in the
control counts as a \emph{false positive}.

\begin{figure}[t]
\centering
\includegraphics[width=\linewidth]{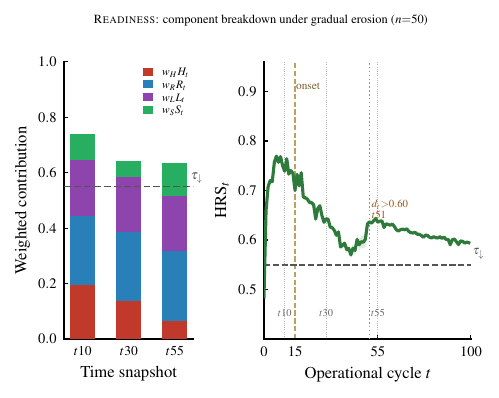}
\caption{HRS component breakdown under gradual sub-threshold erosion
($\Delta\sigma=0.010$\,cycle$^{-1}$, \READ{} condition, $n=50$ seeds;
\textit{computational simulation}).
\textit{Left}: weighted component contributions
($w_H H_t$, $w_R R_t$, $w_L L_t$, $w_S S_t$) at three snapshots.
$H_t$ is the primary driver of composite decline; $R_t$ and $L_t$
remain near ceiling because trust and learning stability are not
directly disrupted by the gradual fatigue--skill shock.
After reversion ($t\approx33$), $S_t$ grows as the system accumulates
tenure in the lower mode, partially compensating the $H_t$ loss and
stabilising the composite above $\tau_{\downarrow}$.
\textit{Right}: \READ{} HRS trajectory with $\tau_{\downarrow}$ and the
cycle at which the individual deskilling guard ($d_t>0.60$) would fire
($t\approx51$) --- well after the composite crosses $\tau_{\downarrow}$
at $t\approx33$, confirming the composite acts before any single guard.}
\label{fig:components}
\end{figure}

All three cap-equipped variants detect deterioration (50/50): the cap
enables reversion. The point of separation is false positives.
$H_t$-only reverts in 50/50 control runs --- a single fatigue signal
cannot distinguish genuine multi-signal erosion from ordinary
fluctuation. Both composite variants (no-hysteresis and \READ{} full)
produce 0/50 false positives: $R_t$, $L_t$, $S_t$ keep $\HRS_t$ above
$\tdown$ during healthy operation, so the composite does not fire
spuriously. The advantage is the composite sensor, not the cap alone.
Figure~\ref{fig:components} decomposes the corresponding HRS trajectory.

\begin{table}[ht]
\centering
\caption{Non-trivial comparators for the gradual-deterioration regime
(software, $n=50$ seeds). All cap-equipped variants detect
deterioration; only the composite eliminates false positives.}
\label{tab:h3comparators}
\scriptsize
\setlength{\tabcolsep}{2pt}
\renewcommand{\arraystretch}{1.15}
\begin{tabular}{@{}
  >{\raggedright\arraybackslash}p{0.46\linewidth}
  >{\centering\arraybackslash}p{0.23\linewidth}
  >{\centering\arraybackslash}p{0.23\linewidth}
  @{}}
\toprule
\textbf{Variant} & \textbf{Detection} & \textbf{False positives} \\
 & {\scriptsize(rev./deterioration)} & {\scriptsize(rev./control)} \\
\midrule
\BASELINE{} (no cap)       & $0/50$ & $0/50$ \\
\GOV{} (guards, no HRS)    & $0/50$ & $0/50$ \\
$H_t$-only (cap)           & $\mathbf{50/50}$ & $50/50$ \\
No-hysteresis (cap)        & $\mathbf{50/50}$ & $\mathbf{0/50}$ \\
\READ{} full (cap)         & $\mathbf{50/50}$ & $\mathbf{0/50}$ \\
\bottomrule
\end{tabular}
\end{table}

\subsection{Component Ablation and Robustness}
\label{sec:results:ablation}

\textbf{Component ablation.}
The headline benchmark tests one perturbation model: gradual
sub-threshold erosion. To identify which HRS components drive detection
when shocks are abrupt rather than gradual, we apply three
direct signal-injection shocks (full data in~\cite{pelechano2026thms_supp},
Section~S1): \textsc{Trust-Drop}, \textsc{Fatigue-Spike}, and
\textsc{Sustained-Degradation}.
\READ{} retains high coverage across all three (50/50, 48/50, 50/50);
\GOV{} stays near zero (1/50, 0/50, 0/50). Two findings clarify which
design elements are responsible.

\emph{Hysteresis}: removing the gap between $\tup$ and $\tdown$ leaves
reversion coverage unchanged. Hysteresis controls escalation stability
--- preventing repeated mode changes near the boundary --- not reversion
triggering.

\emph{Composite vs.\ single signal}: replacing the full HRS with $H_t$
alone preserves most coverage on these shocks, because all three
directly perturb the human-readiness channel. This contrasts with the
gradual-deterioration regime (Section~\ref{sec:results:shocks}), where
$H_t$-only produced 50/50 false positives under the no-deterioration
control: slow multi-signal erosion cannot be distinguished from ordinary
fatigue fluctuation without the trust, stability, and performance
components. Component necessity is therefore perturbation-type
dependent: $H_t$ suffices for acute human-state shocks; the full
composite is required when erosion accumulates across channels.

\textbf{Mode-change frequency.}
A concern with composite governance is that richer sensing could trigger
more frequent mode changes, adding operational churn. The HRS and
hysteresis layer add zero overhead: mean mode-change frequency per
episode was identical under \GOV{} and \READ{}. The hysteresis gap
($k_{\uparrow}=3$ consecutive qualifying cycles before escalation)
prevents oscillation without suppressing necessary reversion.

\subsection{Sensitivity and Stability (H4)}
\label{sec:results:sensitivity}

H4 asks two questions. First: are the HRS's governance decisions correct?
Second: is that correctness robust, or does it depend on a single
precisely tuned configuration?

\textit{Handover accuracy} (HA) measures the proportion of non-trivial
governance decisions --- escalations and reversions --- that are
retrospectively confirmed as warranted by subsequent trajectory outcomes:
an escalation is confirmed if mean task quality over cycles $t{+}1$ to
$t{+}3$ does not fall below quality$_t - 0.02$; a reversion is confirmed
if skill at cycle $t{+}3$ satisfies skill$_{t+3} \geq$ skill$_t - 0.02$
(i.e., the reversion successfully arrested decline).
HA is evaluated under the standard gradual-erosion protocol
($\delta\sigma = 0.010$/cycle, $\delta f = 0.012$/cycle,
$t \in [15,55]$), which provides the realistic signal-to-noise conditions
needed to assess decision quality. A value of 1.00 indicates every
governance decision was confirmed by the subsequent trajectory.

\begin{figure}[t]
\centering
\includegraphics[width=\linewidth]{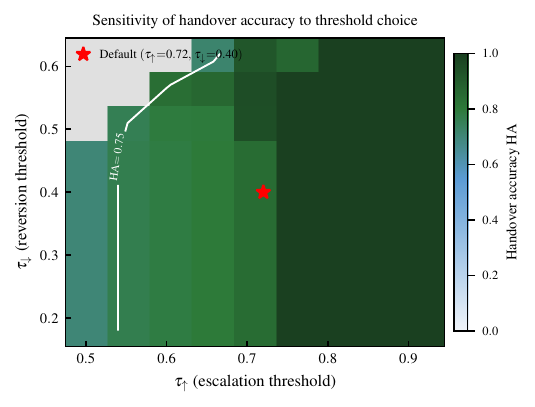}
\caption{Threshold sensitivity sweep: handover accuracy (colour) over the
$(\tup,\tdown)$ parameter grid (9$\times$9, 15 seeds per point,
gradual-erosion protocol). Grey cells denote invalid pairs
($\tdown \geq \tup$). The stable region $\Omega$ ($\mathrm{HA}\geq 0.75$)
covers 91\% of valid pairs; its main boundary is at $\tup \leq 0.50$.
The default ($\tup=0.72$, $\tdown=0.40$, starred) achieves
$\mathrm{HA}=0.85$, well inside $\Omega$.}
\label{fig:sensitivity}
\end{figure}

We swept $\tup \in [0.50,0.92]$ and $\tdown \in [0.18,0.62]$ across a
$9\times 9$ grid (15 seeds per point, gradual-erosion protocol) to map
where governance remains reliable. The stable region $\Omega$ ---
defined by $\mathrm{HA} \geq 0.75$ --- spans the vast majority of the
searched space (91\% of valid threshold pairs; Fig.~\ref{fig:sensitivity}).
The main boundary lies at $\tup \leq 0.50$, where over-aggressive
escalation reduces accuracy ($\mathrm{HA}=0.70$); all tested values of
$\tup \geq 0.55$ achieve $\mathrm{HA} \geq 0.75$ across the full valid
range of $\tdown$. Within $\Omega$, mean $\mathrm{HA} = 0.91 \pm 0.10$.

The default parameters ($\tup=0.72$, $\tdown=0.40$) sit well inside
$\Omega$, achieving $\mathrm{HA} = 0.85$ at the nearest evaluated grid
point. The system is not tuned to a single sweet spot but operates
reliably across a wide range of configurations.

Component weights are similarly robust: $\pm 0.10$ perturbations to
$w_H$, $w_R$, $w_L$ leave $\mathrm{HA} = 1.00$. The exception is $w_S$
(operational performance): reducing it from 0.20 to 0.10 drops
$\mathrm{HA}$ to 0.703. This indicates that $w_S$ shapes the timing and
confidence of mode transitions --- when the system acts, not whether it
detects the need to act.
Full threshold and weight maps appear in~\cite{pelechano2026thms_supp},
Section~S4.

% =============================================================
%  6. DISCUSSION
% =============================================================

\section{Discussion}
\label{sec:discussion}

\textbf{Core claim: complementarity.}
The results establish that hard safety guards and the HRS are not
interchangeable --- they address different failure types across different
time scales. Guards operate on \emph{events}: a single signal crossing a
hard threshold at cycle $t$ triggers immediate reversion, independent of
history. The HRS operates on \emph{trajectories}: it integrates signals
across cycles, requiring no individual threshold to be crossed but
acting when the composite accumulation of fatigue, eroded trust, skill
instability, and deskilling jointly crosses the governance boundary.
This temporal asymmetry reflects a fundamental distinction in failure
types: abrupt events (acute fatigue spike, trust collapse, output
failure streak) demand immediate response, whereas gradual erosion
demands accumulated evidence before action, or the governance layer
would produce excessive false positives
(as the $H_t$-only comparator demonstrates: 50/50 false positives under
no-deterioration conditions). A complete governance architecture requires
both layers. Guard thresholds should be calibrated to the HRS-governed
trajectory, not to an unregulated baseline, since the HRS changes the
signal environment that guards operate in. In domains where structural
constraints (regulatory ceilings, professional floors) already enforce
conservative mode use, the HRS's marginal detection contribution
diminishes; its value then shifts to auditability and pre-positioning
--- keeping the collaboration trajectory documented and recoverable
(confirmed in a radiology screening boundary analysis,
\cite{pelechano2026thms_supp}, Section~S5).

\textbf{Relation to existing frameworks.}
The HRS extends Parasuraman and Riley's~\cite{parasuraman1997humans}
hybrid trigger concept --- which combines multiple signal types to
adapt automation level --- in three directions: it governs \emph{both}
escalation and reversion (not only reversion on failure); it requires
sustained evidence for escalation but reverts immediately, encoding the
asymmetric cost structure of the problem; and it exposes per-component
weights to make each decision auditable. Sheridan's 10-level
LOA taxonomy~\cite{sheridan1978human} describes what modes exist; the
HRS addresses the orthogonal question of \emph{when to move between them
over time} --- the two are complementary, not competing.

\textbf{Non-compensatory properties of the HRS.}
The linear additive form of Eq.~\ref{eq:hrs} raises a legitimate concern:
a dangerously low $H_t$ could in principle be offset by high
$R_t$/$L_t$/$S_t$. Two design features limit this risk. First, $H_t$
itself is multiplicative (Eq.~\ref{eq:H}): any severely degraded factor
collapses the product regardless of the others. Second, the hard guards
(Eq.~\ref{eq:guard}) impose non-compensatory backstops at extreme
conditions ($f_t>0.85$, $\tau_t<0.25$, $d_t>0.60$, $\theta_f=3$).
In the sub-guard range, compensation remains a limitation; a
geometric-mean or min-based composite would eliminate it at the cost of
auditability. The hard guards are the practical defence for worst-case
scenarios, which is precisely why the two layers are designed to
coexist rather than substitute.

\textbf{Operator monitoring and ethical deployment.}
The signal proxies in Table~\ref{tab:signals} --- session length, input
latency, override frequency, confirmation delay --- constitute continuous
operator monitoring, creating ethical obligations a production deployment
must address. Operators should be informed of which signals are collected
and how they feed the decision (\emph{transparency}); autonomous demotion
must be contestable, with a visible account of why a reversion fired and a
channel to dispute it (\emph{contestability}); and monitoring override
frequency may introduce Goodhart-law effects, as operators who know
overrides are penalised may suppress legitimate corrections. These are
design constraints, not secondary concerns, shaping which signals are
collected, how they are used, and who can access the records.

\textbf{The overtrust limitation and calibrated trust.}
Setting $R_t=\tau_t$ treats higher learned trust as higher readiness.
The trust in automation literature \citep{lee2004trust,skitka1999does}
shows that overtrust is as dangerous as undertrust. To evaluate this
risk, we ran a controlled \emph{silent AI degradation} experiment: at
cycle 20 the AI quality drops to 50\% of baseline without any
guard-triggering failure (degraded quality $\approx 0.37$, without
producing a fail streak of length $\theta_f=3$), so trust decays only through natural dynamics. We compare
the original $R_t=\tau_t$ against a calibrated form that penalises
trust--performance mismatch:
\begin{equation}
R_t = \tau_t \cdot \bigl(1 - \mathrm{clip}(|\tau_t - A_t|,\,0,\,1)\bigr),
\label{eq:rtcal}
\end{equation}
where $A_t$ is the mean observed AI quality contribution at cycle $t$
(computable from per-task quality attribution already present in
Table~\ref{tab:signals}). The outcomes are summarised in
Table~\ref{tab:overtrust}.

\begin{table}[ht]
\centering
\caption{Silent AI degradation experiment ($n=50$ seeds, software domain).
AI quality drops to 50\% at cycle 20 for 40 cycles without triggering
guards. False escalations = escalations during the degradation window.}
\label{tab:overtrust}
\small
\renewcommand{\arraystretch}{1.2}
\begin{tabular}{@{} l c c c @{}}
\toprule
\textbf{Variant} & \textbf{False esc.} & \textbf{Reversion} & \textbf{Latency} \\
 & \textbf{(window)} & \textbf{coverage} & (med.\ cyc.) \\
\midrule
\READ{} original ($R_t{=}\tau_t$)
  & $\mathbf{17/50}$ & $50/50$ & $3.0$ \\
\READ{} calibrated (Eq.~\ref{eq:rtcal})
  & $\mathbf{0/50}$ & $50/50$ & $3.0$ \\
\bottomrule
\end{tabular}
\end{table}

With the original $R_t$, the mean HRS \emph{rises} from 0.72 to 0.79
during the degradation window (trust remains elevated while $H_t$,
$L_t$, $S_t$ fluctuate), causing 17/50 seeds to escalate autonomy
precisely when AI performance has collapsed. The calibrated formula
eliminates all 17 false escalations while preserving reversion coverage
(50/50) and latency (3.0 cycles). We therefore recommend the calibrated
form for deployment. Until it is validated against real operator data,
$R_t=\tau_t$ should be used with caution in contexts where AI
overconfidence is plausible.

\subsection*{Threats to Validity}

\textbf{Internal validity.}
All evidence is computational and co-designed with \PLAT{}: the evaluation
cannot fully distinguish ``the HRS is effective'' from ``the HRS is adapted
to \PLAT{}'s mechanics.'' Mitigation: three-domain evaluation and
comparator analysis (Table~\ref{tab:h3comparators}) rule out structural-cap
artefacts; full resolution requires independent platform replication.

\textbf{External validity.}
Trust, fatigue, and skill are modelled parameters, not human
measurements; quantitative calibration to field data remains open. The
LinUCB allocator is a demanding baseline but does not replicate human
decision-making under uncertainty.

\textbf{Construct validity.}
Observable signals operationalise latent constructs; mapping to field
instruments (NASA-TLX \citep{hart1988development}, SAGAT
\citep{endsley1995measurement}) is a prerequisite for deployment.
The \textit{responsible\_score} is a design-coherence check, not an
independent criterion --- non-circular validation requires human studies.

\textbf{Future directions.}
Human-participant validation is the necessary next step, alongside
adaptive weight learning, closed-loop stability analysis, and
multi-operator handovers.

\subsection*{Toward Deployment}

The HRS deployment pathway follows three domain-agnostic steps.
\textbf{Signal proxying}: each component maps to constructs with
established measurement traditions (workload~$\to$~fatigue; reliability
logs~$\to$~trust; performance~$\to$~skill; tenure~$\to$~stability).
\textbf{Shadow calibration}: the HRS runs read-only before activation
while domain experts validate weights and thresholds and the observed
signal distribution informs a one-time hard-guard calibration.
\textbf{Governed go-live}: activate with conservative settings ---
biasing toward \textsc{maintain} and \textsc{revert} --- with explicit
weights a compliance committee can inspect \citep{euaiact2024}. Each
step treats deployment as a governance process, not a technical rollout.

% =============================================================
%  7. CONCLUSION
% =============================================================

\section{Conclusion}
\label{sec:conclusion}

We formalise autonomy handover as a governance problem and show, in
controlled simulation, that a composite readiness gate covers gradual
multi-signal deterioration that threshold guards miss, while preserving
auditability. Hard guards enforce binary threshold conditions; the HRS
adds coverage in the complementary regime where operator readiness erodes
slowly and no individual guard fires. The hysteresis structure ensures
escalation requires sustained evidence while reversion is never delayed.
Three design choices carry transferable lessons: a composite index
integrating signals existing frameworks treat separately; an asymmetric
hysteresis structure mirroring the asymmetric cost structure of the
problem; and explicit design-parameter weights making each transition
decision auditable by component. As human--AI collaboration spreads
across safety-critical and skill-intensive domains, handover-governance
quality will increasingly determine whether these systems are trustworthy.

% =============================================================
%  MANDATORY INCLUSIONS
% =============================================================

\section*{Data Availability Statement}

The data supporting the results reported in this paper are openly
available on Zenodo at \url{https://doi.org/10.5281/zenodo.20798977}.
The deposit contains the supplementary document (PDF, Sections~S1--S5),
all raw simulation outputs as CSV files covering every analysis block
(steady-state metrics, deterioration-response coverage, component
ablation, threshold and weight sensitivity, mode-change frequency, and
the overtrust experiment), and the Python scripts that generate all
figures. Variable definitions are provided in \texttt{CODEBOOK.md}.

\section*{Acknowledgment}

This work was developed with the support of the Spanish Ministry of
Science and Innovation under Project PRODIGIOUS PID2023-146224OB-I00.
The authors used Claude Code (Anthropic) for language editing, clarity
improvements, consistency checks on selected sections, and generating the
\PLAT{} tool's code. Generative AI was not used to produce scientific
claims or conclusions; all AI-assisted content was reviewed and approved
by the authors.

% =============================================================
%  REFERENCES
% =============================================================

\bibliographystyle{IEEEtranN}
\bibliography{references}

\end{document}